\documentclass[letterpaper, 10 pt, conference]{ieeeconf}
\IEEEoverridecommandlockouts
\usepackage{graphicx}
\usepackage{float}
\usepackage{amsmath, amssymb, amsfonts}
\usepackage{cite}
\usepackage{caption}
\usepackage{subcaption}
\usepackage{hyperref}

\usepackage{graphics}
\usepackage{epsfig}
\usepackage{times}
\usepackage[dvipsnames]{xcolor}
\usepackage{booktabs}
\usepackage{multirow}
\usepackage{makecell}
\usepackage{pifont} 
\usepackage{colortbl}

\title{\LARGE \bf
LaViRA: \underline{La}nguage-\underline{Vi}sion-\underline{R}obot \underline{A}ctions Translation for Zero-Shot Vision Language Navigation in Continuous Environments
}

\author{
    \begin{tabular}{c}
        \centerline{Hongyu Ding$^{1,*}$, Ziming Xu$^{1,*}$, YUK TUNG SAMUEL FANG$^{2}$, You Wu$^{1}$, Zixuan Chen$^{1}$,} \\
        \centerline{Jieqi Shi$^{2,\dagger}$, Jing Huo$^{1,\dagger}$, Yifan Zhang$^{3}$, Yang Gao$^{2}$}
    \end{tabular}
    \thanks{\textbf{$^{*}$Equal Contribution}, \textbf{$^{\dagger}$Corresponding Author}}
    \thanks{$^{1}$Hongyu Ding, Ziming Xu, You Wu, Zixuan Chen and Jing Huo are with the School of Computer Science, Nanjing University, China. Emails: {\texttt{\{hongyuding, zimingxu, you\}@smail.nju.edu.cn}}, \texttt{\{chenzx, huojing\}@nju.edu.cn}}
    \thanks{$^{2}$YUK TUNG SAMUEL FANG, Jieqi Shi and Yang Gao are with the School of Intelligence Science and Technology, Nanjing University, China. Emails: {\texttt{231880023@smail.nju.edu.cn, isjieqi@nju.edu.cn, gaoy@nju.edu.cn}}}
    \thanks{$^{3}$Yifan Zhang is with the Institute of Automation, Chinese Academy of Sciences, and the University of Chinese Academy of Sciences, Nanjing, China. Email: {\texttt{yfzhang@nlpr.ia.ac.cn}}}
    \thanks{This work is supported in part by New Generation Artificial Intelligence-National Science and Technology Major Project (2025ZD0122904), National Natural Science Foundation of China (62506153, 62192783, 62276128, 62273347), Jiangsu Science and Technology Major Project (BG2025035), Fundamental Research Funds for the Central Universities (KG202514), Jiangsu Postgraduate Research \& Practice Innovation Program (KYCX25\_0300) and the Collaborative Innovation Center of Novel Software Technology and Industrialization.}
}

\begin{document}

\maketitle
\thispagestyle{empty}
\pagestyle{empty}

\begin{abstract}
Zero-shot Vision-and-Language Navigation in Continuous Environments (VLN-CE) requires an agent to navigate unseen environments based on natural language instructions without any prior training. Current methods face a critical trade-off: either rely on environment-specific waypoint predictors that limit scene generalization, or underutilize the reasoning capabilities of large models during navigation. We introduce LaViRA, a \textit{simple yet effective} zero-shot framework that addresses this dilemma by decomposing action into a coarse-to-fine hierarchy: \textit{Language Action} for high-level planning, \textit{Vision Action} for middle-level perceptual grounding, and \textit{Robot Action} for low-level control.
This modular decomposition allows us to leverage the distinct strengths of different scales of Multimodal Large Language Models (MLLMs) at each stage, creating a system that is powerful in its reasoning, grounding and practical control.
LaViRA significantly outperforms existing state-of-the-art methods on the VLN-CE benchmark, demonstrating superior generalization capabilities in unseen environments,  while maintaining transparency and efficiency for real-world deployment.
Project page: \url{https://robo-lavira.github.io/lavira-zs-vln/}
\end{abstract}

\begin{figure}[ht]
    \centering
    \includegraphics[width=0.95\linewidth]{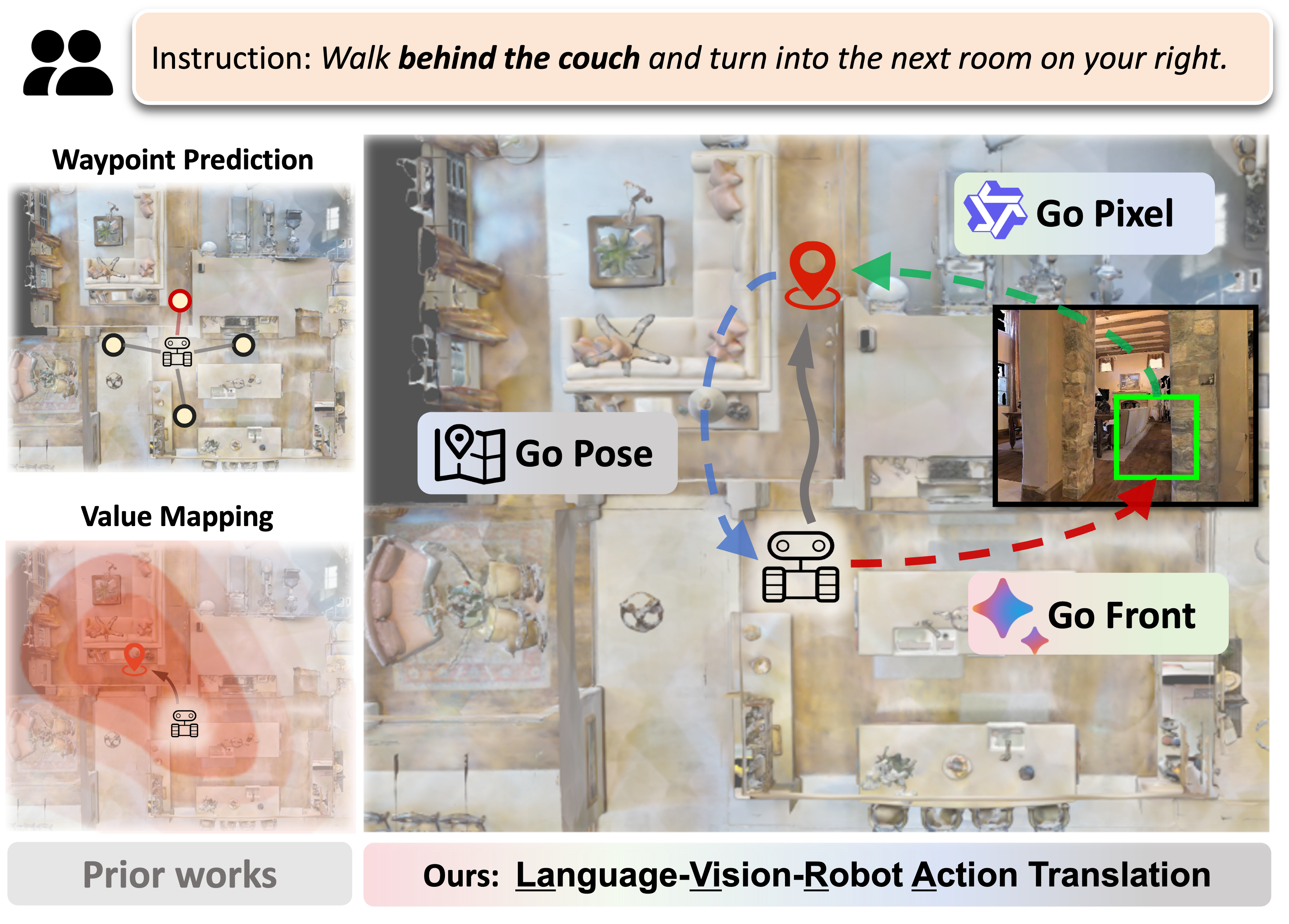}
    \caption{
    Prior methods rely on pre-trained waypoint prediction or value mapping with limited online planning. Our LaViRA framework instead decomposes navigation into language-level planning (``Go Front''), vision-level grounding (``Go Pixel''), and robot-level control (``Go Pose''), fully leveraging MLLMs reasoning for coarse-to-fine decision-making.
    }
    \label{fig:intro}
\end{figure}

\section{Introduction}
Vision-and-Language Navigation (VLN) presents the challenge of grounding natural language instructions within visual observations to enable an embodied agent to navigate through previously unseen environments~\cite{anderson2018vision}.
Early VLN research was primarily conducted in discrete, graph-based settings where navigation is simplified to selecting paths between predefined nodes. To bridge the gap to the real world, Vision-and-Language Navigation in Continuous Environments (VLN-CE)\cite{krantz_vlnce_2020} was introduced, removing the reliance on connectivity graphs and forcing agents to contend with realistic challenges like continuous visual perception and fine-grained motor control.


The recent success of LLMs and MLLMs has inspired zero-shot VLN-CE~\cite{qiao2024opennav, shi2025smartway, chen2025constraint, long2024instructnav}, where agents navigate without environment-specific training. Two main paradigms have emerged.
\textbf{Waypoint prediction with large model reasoning.} These methods combine a pre-trained waypoint predictor with an LLM or MLLM~\cite{qiao2024opennav, shi2025smartway}. The predictor proposes navigable waypoints, and the large model reasons over these candidates to select the next step. This leverages high-level planning but requires separate pre-training and struggles to generalize to unseen environments.
\textbf{Value mapping with vision language models.} An alternative discards the waypoint predictor and uses a vision-language model (e.g., BLIP~\cite{li2023blip2}) to generate a semantic relevance heatmap over the scene. The agent navigates toward the highest-scoring region. While this avoids predictor training, it typically limits powerful large models to offline instruction parsing, underutilizing their reasoning capabilities during online navigation.

These two paradigms reveal a fundamental trade-off. Waypoint-based methods excel at high-level reasoning but are constrained by a separate, inflexible waypoint prediction module. Value-mapping methods are more perceptually grounded but lack dynamic, high-level reasoning during navigation. This leads us to a simple but critical question:

\begin{quote}
    \textit{
    Can we design a purely zero-shot VLN-CE framework that (1) removes the dependency on a pre-trained waypoint predictor and (2) fully harnesses the reasoning abilities of MLLMs for navigation decision making?
    }
\end{quote}

We answer this question with \textbf{LaViRA}: Language-Vision-Robot Action Translation.
Our key insight is to decompose navigation into a coarse-to-fine, multi-stage action space, progressively refined from language to vision to robot control. Grounded in the divide-and-conquer principle for long-horizon decision making under partial observability~\cite{bellman1966dynamic} and hierarchical reinforcement learning theory~\cite{sutton1999between}, this decomposition exponentially reduces the search space at each stage, improving decision efficiency and zero-shot generalization in unseen continuous environments. Rather than requiring a single model to produce low-level controls directly, we allocate each stage to a model scale that best matches its reasoning or perceptual demands, allowing the system to exploit the complementary strengths of different models.

\begin{enumerate}
    \item \textbf{Language Action:} A powerful MLLM acts as a high-level planner, analyzing the instruction, history, and current observation to produce a coarse strategic decision, such as which general direction to head, whether to backtrack, or when to stop.
    \item \textbf{Vision Action:} A smaller, efficient MLLM takes this high-level decision and grounds it in the visual scene, identifying a specific object or region to move towards.
    \item \textbf{Robot Action:} A simple, rule-based controller executes the low-level movement to the identified target.
\end{enumerate}



This hierarchical decomposition offers three key advantages. First, it is \textit{purely zero-shot}, eliminating any need for pre-trained waypoint predictors. Second, it \textit{fully engages MLLM reasoning} across multiple granularities: high-level planning, mid-level perceptual grounding, and low-level control. Third, its modular design provides \textit{transparency and practicality}, allowing seamless adaptation to both simulated and real-world robots.

Our contributions are as follows:
\begin{itemize}
    \item We propose a general action decomposition strategy for zero-shot VLN-CE that separates navigation into language-level planning, vision-level grounding, and robot-level control, enabling flexible integration of reasoning and perception modules.
    \item We instantiate this strategy in LaViRA, a practical Language–Vision–Robot Action framework that leverages multi-scale MLLMs in a fully zero-shot manner.
    \item We achieve state-of-the-art zero-shot performance on the VLN-CE benchmark while maintaining strong effectiveness and efficiency for real-world deployment.
\end{itemize}

\section{Related Work}
\subsection{Vision-and-Language Navigation}
Early research in Vision-and-Language Navigation (VLN)~\cite{anderson2018vision}, where an agent follows instructions to navigate unseen environments, predominantly focused on discrete, graph-based settings~\cite{chang2017matterport3d, qi2020reverie, ku2020room}.
Such environments enable high-level decision-making but fail to capture the continuous control demands of real-world scenarios.
To address this, VLN in Continuous Environments (VLN-CE)~\cite{krantz_vlnce_2020} removes reliance on connectivity graphs and requires agents to perform fine-grained control actions like moving forward, rotating, and avoiding obstacles.
This transition introduces new challenges in scene analysis, generalization, and low-level control.

Numerous VLN methods in both discrete and continuous environments have improved performance through enhanced cross-modal alignment~\cite{hong2021vln, qi2021road}, reinforcement learning~\cite{wang2019reinforced}, data augmentation~\cite{Li2022EnveditEE, fried2018speaker, wang2023scaling, mei2025urbannav}, and map-based representations~\cite{chen2022duet, an2024etpnav, wang2023gridmm}.
However, these learning-based approaches require substantial environment-specific training, which limits their applicability for zero-shot deployment—motivating recent interest in training-free VLN-CE solutions.


\subsection{Zero-Shot VLN with Foundation Models}
The rise of foundation models such as LLMs~\cite{brown2020language, touvron2023llama} and MLLMs~\cite{radford2021clip, li2023blip2, team2023gemini} has driven a new wave of zero-shot VLN-CE research, categorized by how they integrate with robot control.
One dominant paradigm is \textit{waypoint-based navigation}~\cite{qiao2024opennav, shi2025smartway}. These methods use an LLM/MLLM to select waypoints proposed by a pre-trained predictor~\cite{hong_2022_bridging_the_gap}. However, this creates a critical dependency on the predictor, which often fails to generalize to unseen scenes and limits backtracking flexibility~\cite{shi2025smartway}.
Another approach is \textit{heuristic value-mapping}~\cite{chen2025constraint, long2024instructnav}, where a Vision-Language Model (VLM) generates a semantic heatmap to guide the agent. In these frameworks, powerful LLMs are restricted to offline instruction parsing, underutilizing their dynamic reasoning during navigation. Reliance on hard constraints for progress estimation also introduces rigidity in complex scenarios~\cite{chen2025constraint}. Recently, GC-VLN~\cite{yin2025gcvln} models instructions as explicit graph constraints for training-free navigation. While effective for structured reasoning, it still requires offline parsing and lacks flexible backtracking.

Our LaViRA framework bridges the gap between these approaches, introducing a \textit{Language--Vision--Robot Action Translation} that progressively refines action from language-level planning to vision-level grounding to robot-level execution. This design allows MLLMs reasoning to directly influence decisions at multiple granularities, enabling fully zero-shot, interpretable, and adaptable navigation in continuous environments.

\begin{figure*}[ht]
    \centering
    \includegraphics[width=0.95\linewidth]{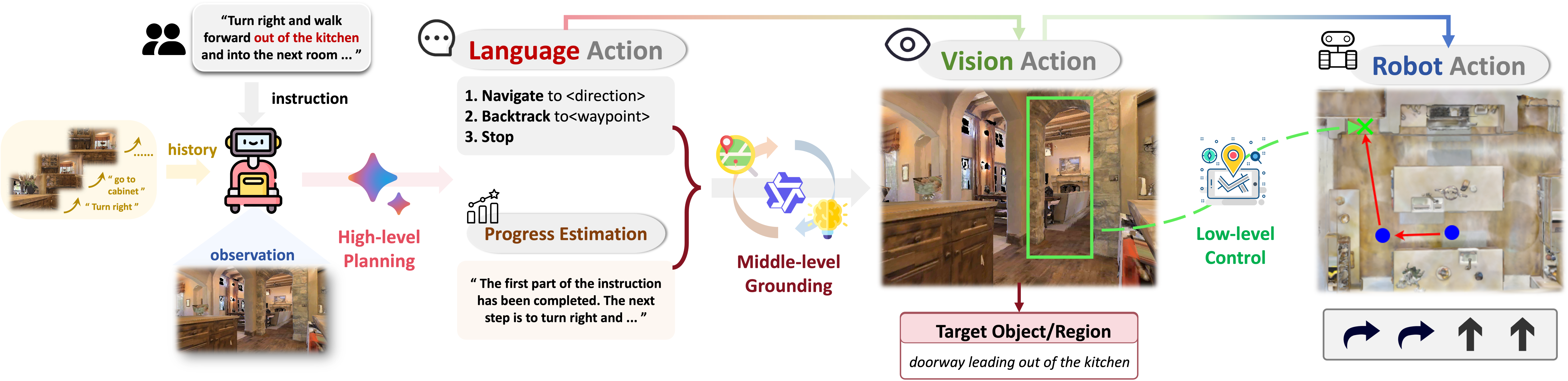}
    \caption{\textbf{The LaViRA Pipeline.} Our framework decomposes navigation into three sequential stages. (1) \textbf{Language Action}: A large MLLM processes the instruction, history, and current observation to generate a high-level plan, deciding whether to move forward, turn, backtrack, or stop. (2) \textbf{Vision Action}: A smaller, more efficient MLLM takes the high-level plan and progress estimation to identify a specific visual target in the chosen direction, outputting its bounding box and description. (3) \textbf{Robot Action}: The target's pixel coordinates are projected onto a global map, and a rule-based controller navigates the robot to the destination. This hierarchical process enables generalized and robust zero-shot vision-language navigation.}
    \label{fig:overview}
\end{figure*}

\subsection{Hierarchical Structure in VLN}
\label{related-work-hierachical}
Hierarchical architectures have emerged as a powerful paradigm for addressing complex, long-horizon navigation tasks. This strategy effectively decomposes the problem into high-level planning and low-level execution, allowing different modules to specialize. Current hierarchical approaches often combine high-level semantic planning with low-level motion control. One common strategy involves dividing the environment into distinct topological regions and having a high-level planner select a sequence of zones to traverse, while a low-level controller navigates within each zone~\cite{liu2023azhp, 2024HOZ}. Other methods learn policies for both the manager and worker through techniques like imitation learning~\cite{2021hcm} or reinforcement learning~\cite{2024Feudal}. A recent example, $NavA^3$~\cite{2025NavA}, uses a global policy to identify target regions and a pre-trained model for fine-grained navigation.

However, a common limitation of these methods is their reliance on extensive, environment-specific training, which restricts generalization and hinders zero-shot deployment. In contrast, LaViRA's Language-Vision-Robot action decomposition is fully zero-shot. By leveraging MLLM reasoning and a rule-based controller, it eliminates the need for any training, which simplifies deployment and enhances modularity.

\section{Proposed Method}
\label{sec:method}

Our approach, LaViRA, introduces a novel framework for zero-shot Vision-and-Language Navigation in Continuous Environments (VLN-CE). The core idea is simple yet effective: we decompose the complex navigation task into a coarse-to-fine hierarchy of actions: \textit{Language Action}, \textit{Vision Action}, and \textit{Robot Action}. This modular decomposition allows us to leverage the distinct strengths of different scales of Multimodal Large Language Models (MLLMs) at each stage, creating a system that is powerful in its reasoning, grounding and practical for real-world deployment, all without requiring any environment-specific training.

The overall pipeline is illustrated in Figure~\ref{fig:overview}. In the VLN-CE task, an agent must follow a natural language instruction $\mathcal{I}$ through an unseen environment. At each timestep $t$, it uses an egocentric observation $I_t$ to choose its next action $\mathcal{A}_t$ from a continuous space. To address this, our method decomposes the navigation process into a sequence of three hierarchical actions: a high-level directional plan (Language Action), the grounding of this plan into a specific visual target (Vision Action), and finally, the low-level movement to reach it (Robot Action). We detail each stage below.

\subsection{Language Action: High-Level Planning}
The first stage of our framework addresses the question: \textit{Where should I generally go next?} To answer this, we employ a powerful, large-scale MLLM (e.g., GPT-4o) that functions as a high-level planner. This model is responsible for interpreting the full context of the navigation task and producing a coarse, directional command.

Specifically, the model receives three types of input:
\begin{itemize}
    \item \textbf{Language Instruction $\mathcal{I}$}: The given natural language instruction provided at the start of the task.
    \item \textbf{Current Observation $\mathcal{O}_t$}: A set of four images corresponding to the agent's front, left, right, and back views $\{I_{front}, I_{left}, I_{right}, I_{back}\}$, providing an informative understanding of the current location.
    \item \textbf{Navigation History $\mathcal{H}_t$}: A structured summary of past observations and actions, formatted as a sequence like $\mathcal{H}_t = \{(\mathcal{O}_0, \mathcal{A}_0), (\mathcal{O}_1, \mathcal{A}_1), \ldots, (\mathcal{O}_{t-1}, \mathcal{A}_{t-1})\}$. This provides crucial context on what has already been accomplished.

\end{itemize}

Given this rich multimodal context, the MLLM performs two tasks simultaneously. First, it generates a \textbf{Progress Estimation $\mathcal{P}_t$}, a natural language assessment of how much of the instruction has been completed. This explicit reasoning step forces the model to track its progress against the overall instruction.
Second, based on its analysis, the model selects a \textbf{Language Action $\mathcal{A}^{lang}_t$} from a discrete set:
\begin{itemize}
    \item \texttt{navigate to <direction>}: Move forward, left, right, or behind.
    \item \texttt{backtrack to <waypoint>}: Return to a previously visited waypoint location.
    \item \texttt{stop}: Terminate the navigation.
\end{itemize}
This process can be formulated as:
\begin{equation}
(\mathcal{A}^{lang}_t, \mathcal{P}_t) = \text{MLLM}_{large}(\mathcal{I}, \mathcal{H}_t, \mathcal{O}_t)
\end{equation}
This stage effectively abstracts the continuous environment into a few high-level choices, allowing the large MLLM to focus its powerful reasoning capabilities on strategic, long-term planning.

\begin{figure*}[ht]
    \centering
    \includegraphics[width=0.95\textwidth]{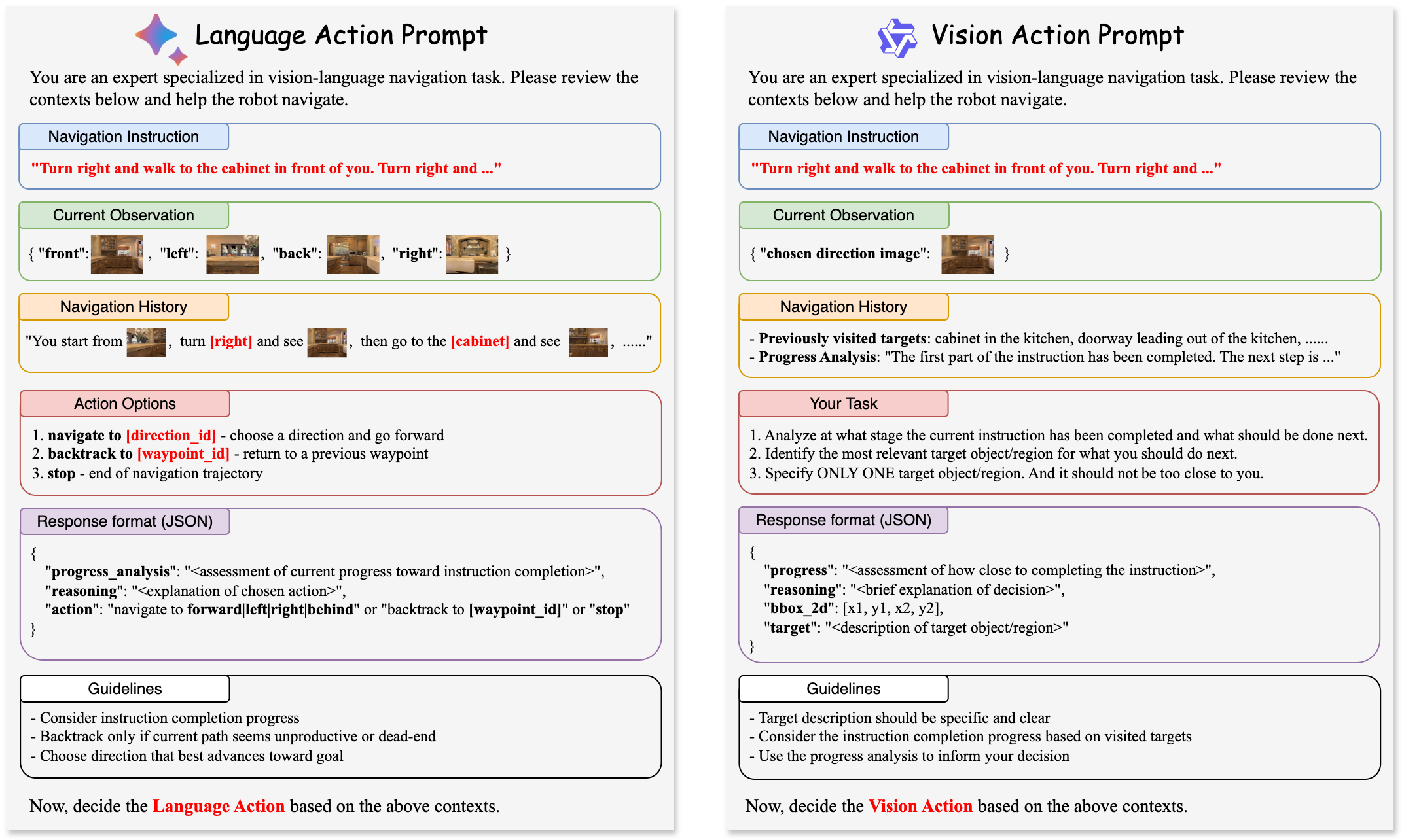}
    \caption{\textbf{Prompts for Language and Vision Actions.} (Left) The prompt for the Language Action model, which takes in full context to decide on a high-level direction. (Right) The prompt for the Vision Action model, which uses the output from the first stage to ground the decision in a specific visual target.}
    \label{fig:prompts}
\end{figure*}

\subsection{Vision Action: Perceptual Grounding}
Once a high-level Language Action is determined, the next stage must answer: \textit{What specific thing should I move towards in that direction?} This is the role of the Vision Action stage, where we ground the abstract plan into a concrete perceptual target.

For this task, we use a smaller, more efficient MLLM (e.g., Qwen2.5-VL-32B). This choice is deliberate: grounding is a more focused perception task that does not require the same extensive world knowledge as high-level planning, making a smaller model more suitable and computationally efficient. As our ablation studies in Section~\ref{model-sec} confirm, pairing a powerful planner with a specialized, efficient grounding model yields optimal performance. The model is prompted with:
\begin{itemize}
    \item \textbf{Language Instruction $\mathcal{I}$}: The original instruction.
    \item \textbf{Progress Estimation $\mathcal{P}_t$}: The text generated by the Language Action model.
    \item \textbf{Chosen Direction Image $I_{dir}$}: The single image corresponding to the direction chosen in the previous stage $\mathcal{A}^{lang}_t$.
\end{itemize}

The model's task is to identify the most relevant object or region in the image that aligns with the next step of the instruction. It outputs a \textbf{Vision Action $\mathcal{A}^{vis}_t$} in a structured format containing a bounding box and its description. This can be expressed as:
\begin{equation}
\mathcal{A}^{vis}_t = \text{MLLM}_{small}(\mathcal{I}, \mathcal{P}_t, I_{dir})
\end{equation}
The output $\mathcal{A}^{vis}_t$ is a dictionary containing:
\begin{itemize}
    \item \textbf{Bounding Box $bbox_{2d}$}: The 2D coordinates \texttt{[x1, y1, x2, y2]} localizing the target.
    \item \textbf{Target Description}: A textual description of the identified object or region.
\end{itemize}
As shown in Figure~\ref{fig:prompts}, the prompt instructs the model to select a target that is not too close, encouraging meaningful progress. This stage effectively translates the high-level plan into a tangible, visually verifiable goal.

\subsection{Robot Action: Low-Level Control}
The final Robot Action stage answers: \textit{How do I physically get there?} It translates the identified visual target into low-level motor commands using a robust, rule-based controller.

\noindent\textbf{Pixel-to-World Projection.}
First, we select the bottom-center pixel of the target's bounding box and project it into a 3D point in the world frame. This involves unprojecting the 2D pixel to the camera's 3D coordinate system using the intrinsic matrix $\mathbf{K}$:
\begin{equation}
    [X_{cam}, Y_{cam}, Z_{cam}]^T = d \cdot \mathbf{K}^{-1} \cdot [u_{target}, v_{target}, 1]^T
\end{equation}
where $d$ is the depth. This 3D point is then transformed from the camera frame to the world frame using the agent's current pose $\mathbf{T}_{agent} = (x_{agent}, z_{agent}, \theta_{agent})$ to yield a target position $\mathbf{p}_{world} = (x_{world}, z_{world})$ on the 2D map:
\begin{equation}
    \begin{bmatrix} x_{world} \\ z_{world} \end{bmatrix} = \begin{bmatrix} x_{agent} \\ z_{agent} \end{bmatrix} + \begin{bmatrix} \cos\theta & -\sin\theta \\ \sin\theta & \cos\theta \end{bmatrix} \begin{bmatrix} Z_{cam} \\ -X_{cam} \end{bmatrix}
\end{equation}

\noindent\textbf{Path Planning and Control.}
Given the target position $\mathbf{p}_{world}$, the agent computes a short-term path using the Fast Marching Method (FMM). A low-level controller then executes this path with local obstacle avoidance. This deterministic final step grounds the reasoning chain in physical action, ensuring interpretability and making the system adaptable to different robot platforms by simply swapping the controller.

\section{Simulation Experiments}
\label{sec:experiments}

We validate LaViRA in simulation to answer three key questions: (1) How does it compare against state-of-the-art methods on the standard VLN-CE benchmark? (2) Does performance support our hypothesis of using different MLLM scales for different decision granularities? (3) What is the contribution of each component in our framework?

\subsection{Experimental Setup}
\noindent\textbf{Environment and Dataset.}
We use the Habitat simulator~\cite{savva2019habitat} with the VLN-CE dataset~\cite{krantz_vlnce_2020}, which extends the R2R benchmark from Matterport3D (MP3D)~\cite{chang2017matterport3d} for continuous navigation. Following recent zero-shot works~\cite{qiao2024opennav, shi2025smartway}, we report results on a standard 100-episode subset from the validation unseen split. An episode is successful if the agent stops within 3 meters of the target.

\noindent\textbf{Evaluation Metrics.}
We use standard VLN metrics: \textit{Navigation Error (NE)}, the final distance to goal; \textit{Success Rate (SR)}, our primary metric for stopping within 3m; \textit{Oracle Success Rate (OSR)}, SR if stopping at the closest point on the path; and \textit{Success rate weighted by Path Length (SPL)}, which penalizes inefficient paths. 
To account for the \textit{inherent stochasticity} in the outputs of MLLMs, we  \textbf{repeat 3 runs} for each experiment over the 100-episode set and report the \textit{mean and standard deviation} for all metrics.

\noindent\textbf{Implementation Details.}
Our zero-shot framework requires no environment-specific training. For the high-level \textit{Language Action} stage, we evaluate two leading MLLMs: Gemini-2.5-Pro and GPT-4o. For the \textit{Vision Action} stage, we primarily use the efficient Qwen2.5-VL-32B, with other models explored in our ablation studies. The agent's observation is composed of posed 640$\times$480 RGB-D images, low-level path planning is executed using the Fast Marching Method (FMM) on a global map constructed from depth observations. All experiments were conducted on 8 NVIDIA RTX 4090 GPUs for parallel evaluation.

\noindent\textbf{Inference Cost.} 
To provide a clear picture of the computational cost, we report the token usage for GPT-4o + Qwen2.5-VL-32B over the 100-episode validation set. 
On average, each trajectory required approximately 32,682 tokens with 7.93 calls for the high-level planner (GPT-4o) and 8,050 tokens with 7.50 calls for the grounding model (Qwen2.5-VL-32B).
Based on current API pricing, the total inference cost is approximately \textbf{\$0.084 USD per episode}.
This highlights the efficiency of our hierarchical design, where an expensive, powerful model is used for high-impact decisions, while a lightweight model handles the low-level perceptual grounding task.

\begin{table}[ht]
\caption{Main results on the VLN-CE benchmark. LaViRA significantly outperforms all previous zero-shot methods. \textbf{Best} and \underline{second-best} zero-shot results are highlighted.}
\label{tab:main_results}
\vspace{-10pt}
\begin{center}
\resizebox{0.9\linewidth}{!}{
\begin{tabular}{l|cccc}
\toprule
\textbf{Method} & \textbf{NE}$\downarrow$ & \textbf{OSR}$\uparrow$ & \textbf{SR}$\uparrow$ & \textbf{SPL}$\uparrow$ \\
\midrule
\rowcolor{black!20}\multicolumn{5}{c}{\textbf{Supervised Learning}} \\
CMA~\cite{hong2022bridging} & 6.92 & 45 & 37 & 32.2 \\
RecBERT~\cite{hong2022bridging} & 5.80 & 57 & 48 & 43.2 \\
ETPNav~\cite{an2024etpnav} & 5.15 & 58 & 52 & 52.2 \\
BEVBert~\cite{an2023bevbert} & 5.13 & 64 & 60 & 53.4 \\
\midrule
\rowcolor{cyan!20}\multicolumn{5}{c}{\textbf{Zero-Shot}} \\
Random & 8.63 & 12 & 2 & 1.5 \\
NavGPT-CE~\cite{Long2023DiscussBM} & 8.37 & 26.9 & 16.3 & 10.2 \\
DiscussNav-CE~\cite{Long2023DiscussBM} & 7.77 & 15 & 11 & 10.5 \\
MapGPT-CE~\cite{chen2024mapgpt} & 8.16 & 21 & 7 & 5.0 \\
Open-Nav~\cite{qiao2024opennav} & 6.70 & 23 & 19 & 16.1 \\
SmartWay~\cite{shi2025smartway} & 7.11 & \textbf{51} & 29 & 22.5 \\
InstructNav~\cite{long2024instructnav} & 6.89 & 47 & 31 & \underline{24.0} \\
CA-Nav~\cite{chen2025constraint} & 7.58 & 48.0 & 25.3 & 10.8 \\
GC-VLN~\cite{yin2025gcvln} & 7.30 & 41.8 & 33.6 & 16.3 \\
\midrule
\textbf{LaViRA(GPT-4o)} & \textbf{6.43}$\pm$0.28 & 43.3$\pm$3.2 & \underline{36.0}$\pm$1.7 & \textbf{28.3}$\pm$0.8 \\
\textbf{LaViRA(Gemini-2.5-Pro)} & \underline{6.54}$\pm$0.27 & \underline{48.7}$\pm$2.1 & \textbf{38.3}$\pm$0.6 & \textbf{28.3}$\pm$0.9 \\

\bottomrule
\end{tabular}}
\end{center}
\vspace{-15pt}
\end{table}



\subsection{Main Results}
We compare LaViRA with existing methods on the VLN-CE benchmark. As shown in Table~\ref{tab:main_results}, our method sets a new state-of-the-art for zero-shot VLN-CE.

The Gemini-2.5-Pro variant achieves SR of 38.3\% and SPL of 28.3\%, improving over the prior best zero-shot method InstructNav~\cite{long2024instructnav} by 7.3 points in SR and 4.3 points in SPL.
Low standard deviations across runs highlight the framework’s robustness and stability, a key advantage for real-world applications.
Notably, LaViRA surpasses supervised methods in SR, demonstrating the power of our hierarchical decomposition and MLLM reasoning. The GPT-4o variant also delivers strong, stable results. These findings validate our central hypothesis: decomposing the navigation task and leveraging advanced MLLM reasoning enables superior zero-shot performance without any pre-trained waypoint predictor.

\subsection{Ablation Studies}
We performed a series of ablation studies to analyze LaViRA's performance and quantify the contribution of its core components. Our baseline for these studies is LaViRA (GPT-4o + Qwen2.5-VL-32B). Although the Gemini-2.5-Pro variant delivered superior performance, we used the GPT-4o variant for ablations due to documented stability issues with the Gemini-2.5-Pro API during our experiments, which could have compromised the consistency of iterative testing. The GPT-4o model provided the necessary reliability for a rigorous and reproducible analysis.

\noindent\textbf{Model Selection.}
\label{model-sec}
Our framework is model-agnostic, allowing flexible combinations of MLLMs. As shown in Table~\ref{tab:ablation_model}, our experiments confirm that pairing models of different scales according to task granularity is crucial. Using a powerful MLLM (e.g., GPT-4o) for high-level \textit{Language Action} (LA) is key; replacing it with the smaller Qwen2.5-VL-72B leads to a 7.0-point drop in SPL. For the more focused \textit{Vision Action} (VA) stage, an efficient model like Qwen2.5-VL-32B proves highly effective. This aligns with the principles of hierarchical VLN discussed in our related work~\ref{related-work-hierachical}, where specialized modules handle different levels of the task. 
Interestingly, using a powerful model like GPT-4o for both stages significantly degrades performance (SPL drops from 28.3\% to 16.8\%). This suggests that simply using the largest model is not optimal, the best results—pairing a top-tier planning model with an efficient grounding model—validate our core hypothesis and demonstrate a key advantage of our zero-shot hierarchical approach.

\begin{table}[t]
\caption{Ablation on model selection. Performance is maximized when a powerful MLLM handles high-level Language Actions (LAM) and an efficient MLLM handles focused Vision Actions (VAM). Qwen-32B and Qwen-72B are short for Qwen2.5-VL-32B and Qwen2.5-VL-72B.}
\label{tab:ablation_model}
\vspace{-10pt}
\begin{center}
\resizebox{\linewidth}{!}{
\begin{tabular}{ll|cccc}
\toprule
\textbf{LAM} & \textbf{VAM} & \textbf{NE}$\downarrow$ & \textbf{OSR}$\uparrow$ & \textbf{SR}$\uparrow$ & \textbf{SPL}$\uparrow$ \\
\midrule
Qwen-32B & Qwen-32B & 8.04$\pm$0.28 & 25.3$\pm$2.1 & 19.7$\pm$3.5 & 14.3$\pm$2.6 \\
Qwen-72B & Qwen-32B & 7.18$\pm$0.18 & 33.3$\pm$2.1 & 27.7$\pm$2.3 & 21.3$\pm$2.7 \\
\textbf{GPT-4o} & \textbf{Qwen-32B} & \textbf{6.43$\pm$0.28} & \textbf{43.3$\pm$3.2} & \textbf{36.0$\pm$1.7} & \textbf{28.3$\pm$0.8} \\
GPT-4o & Qwen-72B & 6.78$\pm$0.11 & 39.3$\pm$2.1 & 32.3$\pm$1.2 & 23.8$\pm$0.7 \\
GPT-4o & GPT-4o & 7.47$\pm$0.24 & 26.0$\pm$7.5 & 20.7$\pm$5.7 & 16.8$\pm$4.9 \\
\bottomrule
\end{tabular}}
\end{center}
\vspace{-15pt}
\end{table}

\begin{table}[t]
\caption{Ablation on framework design. A three-stage pipeline, rich visual history, and flexible backtracking are all crucial for robust navigation.}
\label{tab:ablation_design}
\vspace{-10pt}
\begin{center}
\resizebox{0.95\linewidth}{!}{
\begin{tabular}{l|cccc}
\toprule
\textbf{Configuration} & \textbf{NE}$\downarrow$ & \textbf{OSR}$\uparrow$ & \textbf{SR}$\uparrow$ & \textbf{SPL}$\uparrow$ \\
\midrule
\textbf{LaViRA (Full)} & \textbf{6.43$\pm$0.28} & \textbf{43.3$\pm$3.2} & \textbf{36.0$\pm$1.7} & \textbf{28.3$\pm$0.8} \\
\midrule
\rowcolor{red!20}\multicolumn{5}{c}{\textbf{Framework Decomposition}} \\
w/o LA & 8.94$\pm$0.53 & 13.0$\pm$3.0 & 6.7$\pm$0.6 & 4.4$\pm$1.2 \\
w/o VA & 7.28$\pm$0.23 & 34.0$\pm$4.0 & 23.0$\pm$5.2 & 13.9$\pm$3.4 \\
w/o LA+VA & 9.21$\pm$0.11 & 1.7$\pm$0.6 & 0.0$\pm$0.0 & 0.0$\pm$0.0 \\
\midrule
\rowcolor{orange!20}\multicolumn{5}{c}{\textbf{History Representation}} \\
text obs. + act. & 6.99$\pm$0.32 & 37.7$\pm$5.5 & 31.0$\pm$3.6 & 23.0$\pm$2.8 \\
only act. & 6.54$\pm$0.11 & 38.3$\pm$1.2 & 32.7$\pm$2.1 & 25.1$\pm$1.8 \\
only visual obs. & 6.64$\pm$0.23 & 36.0$\pm$5.6 & 29.3$\pm$3.5 & 22.5$\pm$2.2 \\
only text obs. & 6.96$\pm$0.24 & 35.7$\pm$4.0 & 27.7$\pm$2.5 & 21.8$\pm$1.8 \\
w/o history & 6.90$\pm$0.46 & 36.3$\pm$7.0 & 27.0$\pm$5.6 & 19.4$\pm$7.3 \\
\midrule
\rowcolor{yellow!20}\multicolumn{5}{c}{\textbf{Backtracking Mechanism}} \\
w/o backtrack & 6.92$\pm$0.32 & 42.0$\pm$3.0 & 30.0$\pm$4.4 & 22.2$\pm$4.0 \\
last waypoint only & 6.65$\pm$0.16 & 41.7$\pm$6.0 & 31.7$\pm$1.5 & 23.5$\pm$1.0 \\
\rowcolor{green!20}\multicolumn{5}{c}{\textbf{Pixel Point Selection}} \\
direct output point & 7.00$\pm$0.84 & 45.3$\pm$3.1 & 31.0$\pm$5.6 & 19.4$\pm$5.3 \\
bbox median depth & 6.73$\pm$0.09 & 51.0$\pm$2.6 & 34.0$\pm$1.0 & 21.8$\pm$1.2 \\
\bottomrule
\end{tabular}}
\end{center}
\vspace{-15pt}
\end{table}

\noindent\textbf{Framework Design Choices.}
We conducted further ablations on key design choices, as shown in Table~\ref{tab:ablation_design}.
First, we ablated the \textbf{hierarchical structure}. An end-to-end baseline (``w/o LA+VA'') fails with 0\% SPL. Removing the high-level planner (``w/o LA'') yields 4.4\% SPL, while removing the perceptual grounding module (``w/o VA'') achieves 13.9\% SPL. Our full framework outperforms the latter by 14.4 points, confirming the effectiveness of coarse-to-fine decomposition.
Next, we analyzed \textbf{history representation}. Replacing visual history with textual descriptions drops SPL by 5.3 points to 23.0\%. Using only past actions (25.1\%) or only textual observations (21.8\%) also underperforms, showing that combining raw visual observations and actions is most effective.
We then evaluated the \textbf{backtracking mechanism}. Disabling it causes a 6.1-point SPL drop. Restricting backtracking to the last waypoint only remains 4.8 points worse than our flexible strategy.
Finally, we ablated \textbf{pixel point selection}. Directly outputting pixel coordinates achieves 19.4\% SPL, which is 8.9 points lower than our bounding-box + bottom-center heuristic approach.

\subsection{Qualitative Analysis}
\begin{figure*}[!ht]
    \centering
    \includegraphics[width=0.95\linewidth]{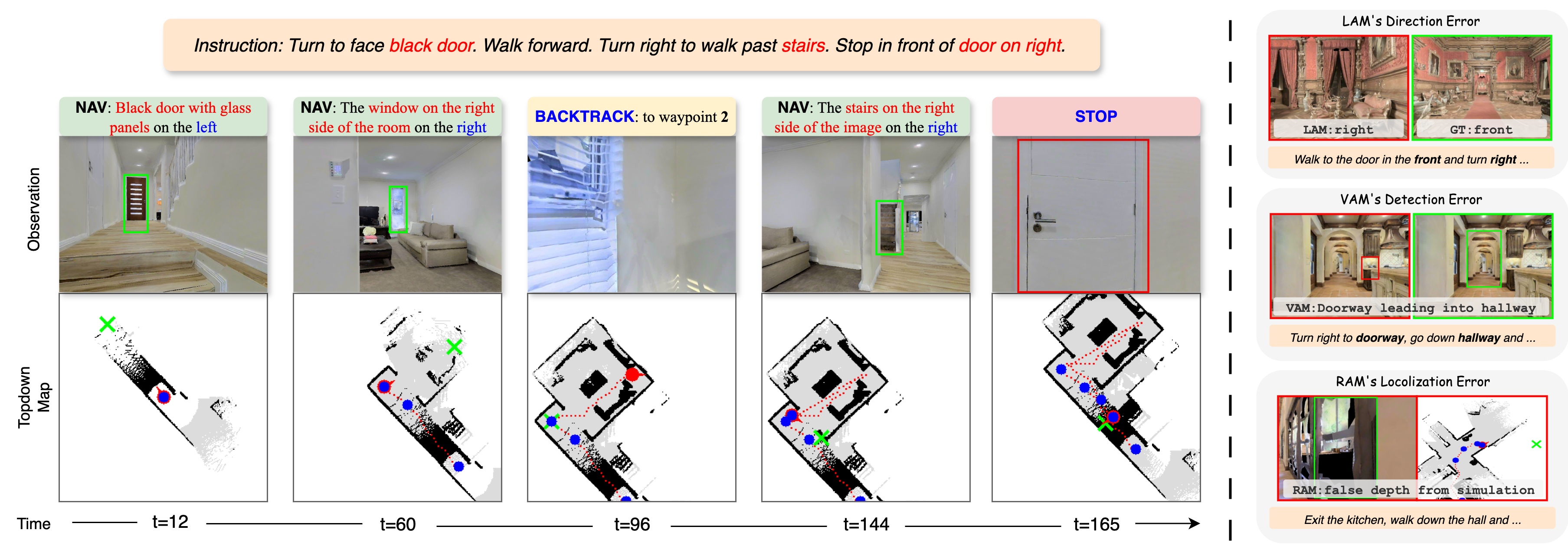}
    \caption{\textbf{Visualization examples.} 
    (Left) Navigation visualization:  
    Language Action outputs are in blue text. Vision Action outputs bounding boxes in green and target descriptions in red. The robot’s pose is a red arrow, history waypoints are blue dots, and the next target position is marked by a green cross.
    (Right) Failure cases visualization:  
    Language Action misjudges direction due to ambiguous instructions; Vision Action selects the wrong region despite correct target description; simulation reconstruction errors cause incorrect target localization.
    }
    \label{fig:vis-all}
\end{figure*}

To offer qualitative insights into LaViRA's decision-making, Figure~\ref{fig:vis-all} shows a successful navigation run and common failures. In the success case, LaViRA demonstrates its coarse-to-fine approach: it first makes a high-level directional choice (``navigate left''), then grounds this in a visual landmark (``Black door with glass panels''), and finally selects a precise waypoint. This hierarchical process validates our design. 


The failure cases illustrate three common errors: (1) A Language Action error from ambiguous instructions, e.g., failing to identify the correct door when multiple doors are visible. This can be mitigated by explicit sub-instruction decomposition during prompting. (2) A Vision Action error where the correct object is grounded to the wrong image region. Although VAM excels at bounding-box grounding, it sometimes fails on larger areas like ``hallway'' or ``living room''. Integrating open-vocabulary segmentation (e.g., SAM 2) as post-processing can address this. (3) A simulation-induced depth error where transparent objects (e.g., windows in Habitat) lack depth values, leading to incorrect 3D projection. In real-world deployment, this can be resolved via LiDAR or stereo depth fusion.

\begin{figure*}[ht]
    \centering
    \includegraphics[width=0.85\linewidth]{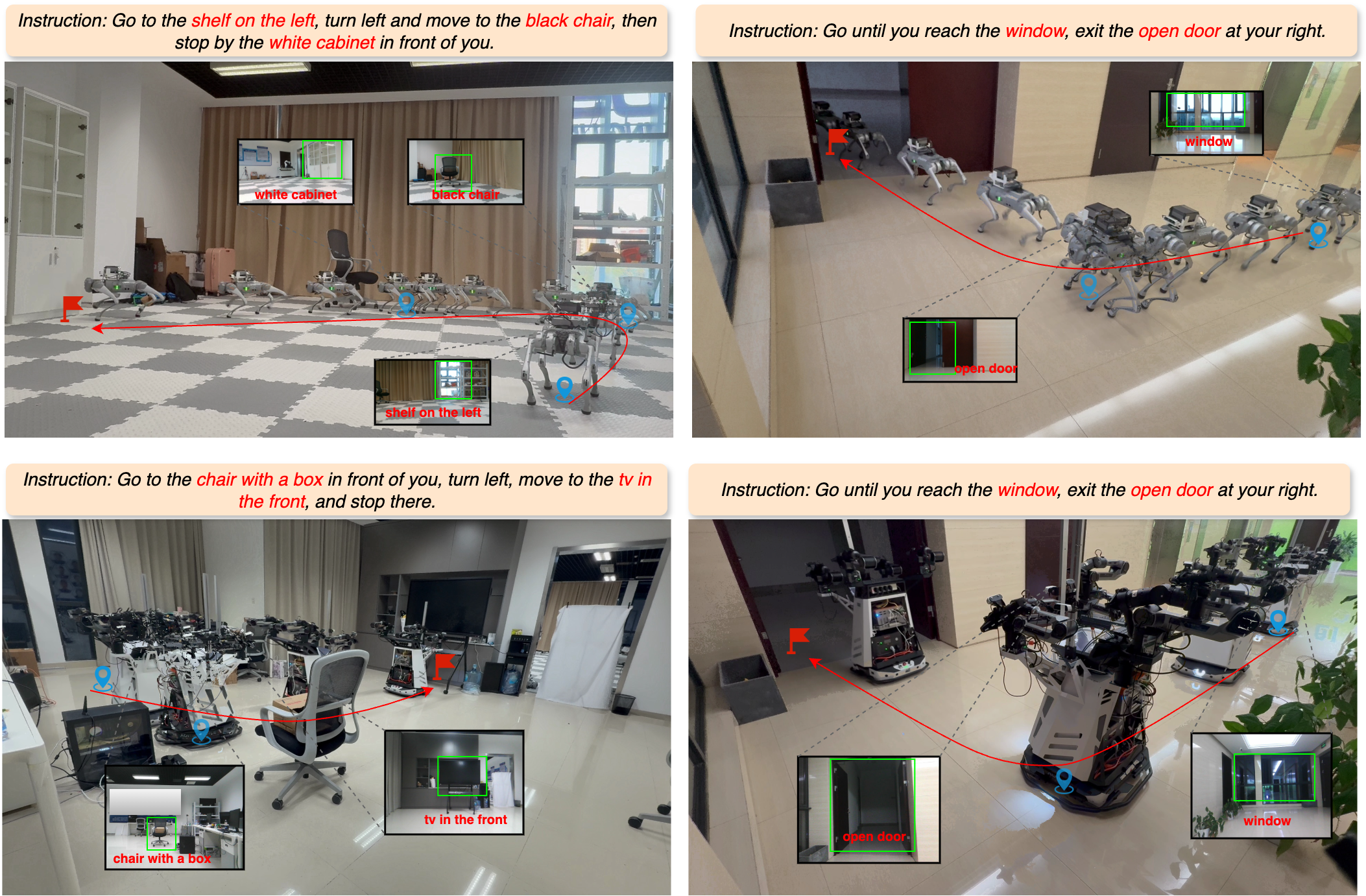}
    \caption{\textbf{Real-world experiment examples.} LaViRA guides a Unitree Go1 quadruped (top) and an Agilex Cobot Magic wheeled robot (bottom) in an office. The visualization shows the third-person view of the robot's trajectory alongside the agent's ego view and targets, demonstrating successful real-world adaptation on diverse platforms.}
    \label{fig:real-world}
\end{figure*}

\section{Real-World Experiments}
To validate LaViRA's practicality beyond simulation, we deployed it on two distinct real-world robots: a Unitree Go1 quadruped and an Agilex Cobot Magic wheeled platform. These experiments tested the framework's sim-to-real transferability, requiring only the replacement of the low-level robot controller.
The Unitree Go1 was equipped with a Jetson Orin NX and an Intel RealSense D435i camera, using its native velocity controller to navigate. The Agilex Cobot Magic platform used a chest-mounted Orbbec Dabai camera and a 2-DOF velocity controller for its mobile base. In both deployments, the onboard computers called the respective MLLM APIs for Language and Vision Actions, transmitting the resulting target pixel coordinates to the robot's navigation system.

As shown in Figure~\ref{fig:real-world}, both platforms successfully executed navigation tasks in complex indoor office environments. The supplementary video further demonstrates consistent performance across multiple cases on each robot, highlighting the framework's strong sim-to-real transferability with only platform-specific low-level controller replacement. These qualitative results confirm that LaViRA's hierarchical reasoning generalizes effectively from simulation to physical hardware without any training.

\section{Conclusion}

We introduce LaViRA, a hierarchical framework for zero-shot Vision-and-Language Navigation in continuous environments. By decomposing navigation into a coarse-to-fine action hierarchy, LaViRA eliminates pre-trained waypoint predictors and fully leverages multi-granularity MLLM reasoning. Experiments show that pairing a powerful MLLM for high-level planning with an efficient one for perceptual grounding achieves state-of-the-art performance on the VLN-CE benchmark, significantly outperforming existing zero-shot methods.

Despite these results, LaViRA relies on proprietary MLLMs, raising concerns about API latency, cost volatility, stability, and limited control. Its performance ceiling is bounded by off-the-shelf models, as seen in failures on ambiguous instructions and large-area grounding. Real-world deployment faces sensor noise, dynamic obstacles, varying illumination, and long-term drift. Future work will distill the pipeline into open-source MLLMs, add caching and adaptation mechanisms to reduce latency and cost, and integrate robust perception for reliable long-term deployment.

\bibliographystyle{IEEEtran}
\bibliography{refs}

\end{document}